\documentclass[runningheads]{llncs}
\usepackage[T1]{fontenc}
\usepackage{graphicx}
\usepackage{amsmath}
\usepackage{amsfonts}
\usepackage{subfigure}
\usepackage{multirow}
\usepackage{sidecap}
\usepackage{booktabs}

\sidecaptionvpos{figure}{c}

\begin{document}
\title{Exploring Attention Map Reuse for Efficient Transformer Neural Networks}
\titlerunning{Exploring Attention Map Reuse for Efficient Transformers}

\author{
Kyuhong Shim\inst{1}\orcidID{0000-0002-0123-3100} \and
Jungwook Choi\inst{2}\orcidID{0000-0002-5691-4771} \and
Wonyong Sung\inst{1,2}\orcidID{0000-0001-8801-210X}
}
\authorrunning{K. Shim, J. Choi, and W. Sung}

\institute{
Dept. of Electrical and Computer Engineering, Seoul National University, Korea \\
\email{skhu20@snu.ac.kr, wysung@snu.ac.kr} \and
Dept. of Electrical Engineering, Hanyang University, Korea \\
\email{choij@hanyang.ac.kr}
}


\maketitle

\begin{abstract}
Transformer-based deep neural networks have achieved great success in various sequence applications due to their powerful ability to model long-range dependency.
The key module of Transformer is self-attention (SA) which extracts features from the entire sequence regardless of the distance between positions.
Although SA helps Transformer performs particularly well on long-range tasks, SA requires quadratic computation and memory complexity with the input sequence length.
Recently, attention map reuse, which groups multiple SA layers to share one attention map, has been proposed and achieved significant speedup for speech recognition models.
In this paper, we provide a comprehensive study on attention map reuse focusing on its ability to accelerate inference.
We compare the method with other SA compression techniques and conduct a breakdown analysis of its advantages for a long sequence.
We demonstrate the effectiveness of attention map reuse by measuring the latency on both CPU and GPU platforms.

\end{abstract}

\section{Introduction}\label{sec:introduction}

The ability to learn long-range dependency is essential for various sequence processing tasks such as language modeling, machine translation, text summarizing, question answering, and speech recognition.
Deep neural networks (DNNs) have been achieved great success in these complex sequence tasks over traditional handcrafted and rule-based techniques.
DNN architectures can be characterized by how the feature extraction mechanism incorporates past or future information.
For example, recurrent neural networks (RNNs) such as LSTM~\cite{lstm} encode the entire previous sequence into a single feature vector, which is beneficial for the compact implementation.
However, it causes a loss of long-range information since feature representation is restricted to a single vector.
In contrast, Transformer~\cite{transformer} models directly access the entire sequence, therefore they are much more advantageous for long-range dependency modeling.
Transformer models have demonstrated excellent performance over RNNs and become the universal choice for most sequence processing applications.

However, Transformer models suffer from quadratic computation and memory complexity to calculate the relationship between every pair of locations.
More precisely, the self-attention (SA) module, one of two submodules in Transformer, utilizes an attention mechanism to process a sequence of length $T$ at once.
The attention mechanism computes the correlation between length $T$ features by the matrix multiplication of the feature matrix ($T\times d$) and its transposed form ($d\times T$), where $d$ is the feature vector dimension.
This $O(T^2)$ quadratic hardware cost is especially problematic when deploying Transformer in practice, especially when the input sequence length is very long.
For example, in speech recognition and language modeling, the length of the input sequence is very long and long-range dependency is crucial for accurate prediction.
Language modeling models usually take about 256 to 512 previous words as input to predict the next word, and speech recognition models often process much more frames (750 frames for 30 seconds) to transcribe the given utterance.
Although these two tasks are core building blocks of many applications, RNN models are still practically favorable because of the heavy computational cost of the Transformer models for resource-limited devices such as mobile and embedded systems.

Considering that a Transformer model is composed of multiple Transformer layers, and each layer consists of multiple (attention) heads that exploit different all-to-all relationships, the complexity increases proportionally to the number of SA heads ($H$) and SA layers ($L$). 
Various architectural modifications have been proposed to reduce the heavy computation of SA, where the studies can be mainly categorized into two groups~\cite{surveyefficient}.
The first group (Section~\ref{ssec:related_attention}) focuses on the output of the attention mechanism, called attention map $A\in\mathbb{R}^{T\times T}$, whereas the second group (Section~\ref{ssec:related_remove}) focuses on reducing the number of $L$ and $H$.
The first group includes 1) computing only a few elements (sparsely) based on patterns or importance~\cite{longformer,sparsetransformer}, and 2) approximating $A$ by low-rank factorization, clustering, or kernelization~\cite{linformer,routingtransformer,performer}.
However, the aforementioned methods cannot utilize the full capability of the modern parallel processing hardware such as graphics processing unit (GPU), digital signal processor (DSP), and neural processing unit (NPU)~\cite{qi2021accelerating,li2020efficient}, due to their unstructured computation savings.
For example, selectively computing a few important elements may produce a random-like access pattern that depends on the input.
Clustering-based methods also require additional K-means or locality-sensitive hashing to group similar ones for a more accurate approximation.
On the other hand, the second group reduces the effective number of attention map computations by pruning out SA heads or SA layers.
These approaches are much more hardware-friendly because the removal of a large computation block (e.g., attention head, attention layer) is very structured and predictable.

Recently, \textbf{attention map reuse} has been proposed for various applications, such as language modeling~\cite{lazyformer}, machine translation~\cite{xiao2019sharing}, and speech recognition~\cite{attention-reuse}.
The key idea of the method is to reuse the attention map of $\ell\text{-th}$ layer $A^\ell$ for multiple consecutive layers, $(\ell+1)\text{-th}$ to $(\ell+M)\text{-th}$ layer, therefore reducing the effective number of SA computation from $L$ to $L/M$.
This architectural change is highly structured and easy to implement on modern hardware platforms.
Especially, for speech recognition, we showed that attention map reuse can be adopted without much degradation of recognition performance~\cite{attention-reuse}.
The paper discovered that the reason behind the success of this method is that SA blocks in successive layers perform a similar role for speech recognition and can be merged. 
However, the analysis on attention map reuse was mainly focused on their behaviors; not much discussion was provided on how the speedup is achieved in terms of the actual implementation.

In this paper, we provide a deeper understanding of attention map reuse in the case of speech recognition.
In Section 2, we briefly introduce Transformer and SA architecture and compare the previous SA compression methods with attention map reuse.
In Section 3, we analyze the effect of each component of a Transformer model for speech recognition.
In Section 4, we evaluate the latency savings of attention map reuse on various configurations and platforms.
The results confirm that attention map reuse is a promising inference speedup technique for Transformer-based long-range sequence processing.
\section{Background and Related Work}\label{sec:background}

\begin{figure*}[t]
    \centering
    \includegraphics[width=0.84\linewidth]{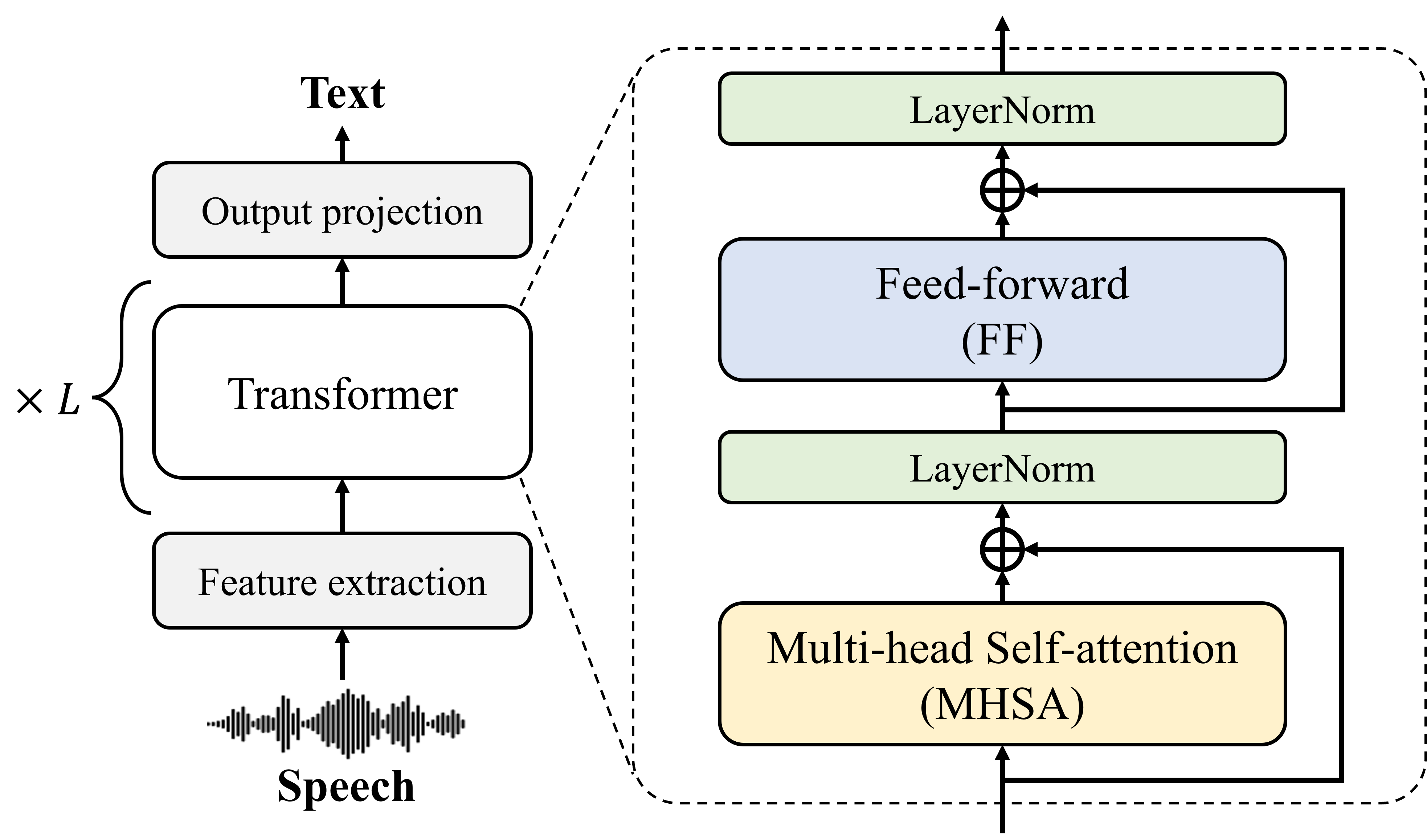}
    \caption{Illustration of the Transformer-based model which is a stack of $L$ Transformer layers.
    For speech recognition, input is human speech and output is a transcribed sentence.
    Note that every frame is processed together without recurrence.
    }
    \label{fig:transformer}
\end{figure*}

\subsection{Transformer and Self Attention}\label{ssec:sa}

\begin{figure*}[t]
    \centering
    \includegraphics[width=0.92\linewidth]{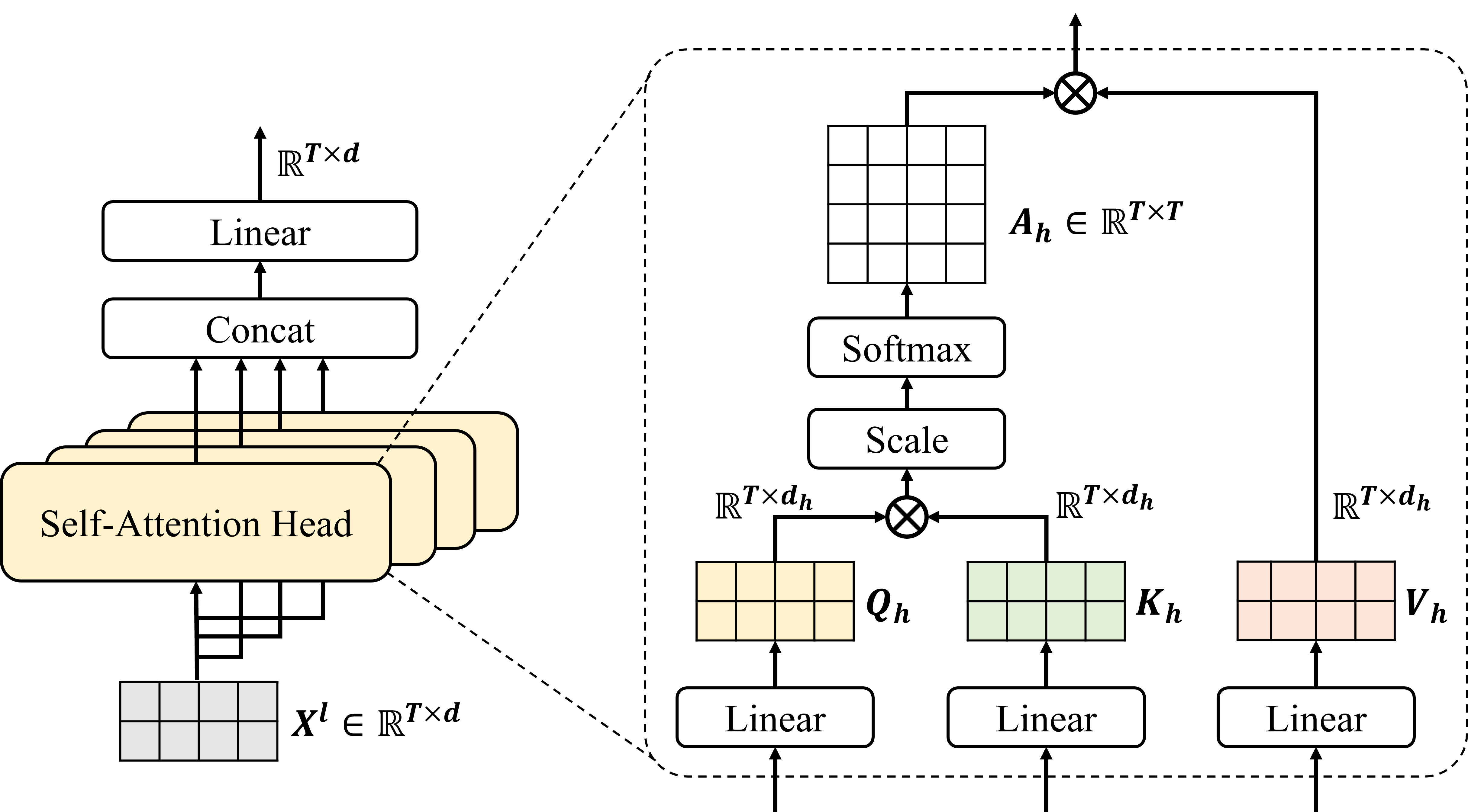}
    \caption{Illustration of the MHSA submodule.
    The computation flow is identical for each attention head.
    }
    \label{fig:mhsa}
\end{figure*}

We briefly introduce the components of a Transformer layer.
A Transformer layer is composed of two submodules: 1) multi-head self-attention (MHSA) and 2) feed-forward (FF).
Figure~\ref{fig:transformer} illustrates the overall architecture, and Figure~\ref{fig:mhsa} visualizes the internal structure of MHSA submodule.
Let the input of a layer be a sequence of $T$ tokens\footnote{We exploit the term `token' as a common concept for both natural language processing and speech recognition. Each token represents a (sub-)word feature and a speech frame feature, respectively.}.
For the input $X=\{x_1, x_2, ...x_T\}$ where $X\in\mathbb{R}^{T\times d}$, SA for the $h\text{-th}$ head (total $H$ heads) starts with three linear projections as:
\begin{equation}
    Q_h,K_h,V_h = XW_{Q_h,K_h,V_h} + b_{Q_h,K_h,V_h} \quad (Q_h,K_h,V_h\in\mathbb{R}^{T\times d_h})
\end{equation}
where $Q,K,V$ indicates the query, key, value, and $W\in\mathbb{R}^{d\times d_h}, b\in\mathbb{R}^{1\times d_h}$ are weight and bias parameters.
$d_h = d/H$ is the feature dimension for each attention head.
The attention map $A_h$ for the $h\text{-th}$ head is then computed as a scaled dot-product of query and key
\begin{equation}
    A_h = \text{Softmax}(\frac{Q_h K_h^T}{\sqrt{d_h}}). \quad (A\in\mathbb{R}^{T\times T})
\end{equation}
After the softmax operation, each row of $A_h$ becomes a probability distribution of a length $T$.
Intuitively, the $(i,j)\text{-th}$ element of the attention map represents how much $j\text{-th}$ token contributes to $i\text{-th}$ token.
Then, the outputs of each attention head are concatenated and followed by another linear projection:
\begin{align}
    \text{SA}_h(X) &= A_h V_h \\
    \text{MHSA}(X) &= \text{Concat}(\text{SA}_1, ... \text{SA}_h)W_O + b_O.  \label{eq:mhsa}
\end{align}
By exploiting multiple attention heads, Transformer can extract diverse relationships between tokens in a single layer.
For example, one head may focus on the syntactic connections while the other head focuses on specific words.

The FF submodule is a stack of two linear projections with an intermediate non-linear activation function:
\begin{equation}
    \text{FF}(X) = \big( \phi(XW_{1} + b_{1}) \big)W_{2} + b_{2}
\end{equation}
where $W_1 \in\mathbb{R}^{d\times 4d}$, $W_2 \in\mathbb{R}^{4d\times d}$, $b_1 \in\mathbb{R}^{4d}$ and $b_2 \in\mathbb{R}^{d}$ are parameters, and function $\phi$ can be ReLU, Swish, GELU, etc.
Finally, the output of $\ell\text{-th}$ Transformer layer is formulated as below
\begin{align}
    Z^{\ell} &= \text{LN}(\text{MHSA}(X^\ell) + X^\ell) \\
    X^{\ell+1} &= \text{LN}(\text{FF}(Z^\ell) + Z^\ell).
\end{align}
where $Z$ is the intermediate term and $\text{LN}$ indicates the layer normalization~\cite{layernorm}.
There exist a residual connection that adds each submodule's input and output.
As $T$ increases, the cost of MHSA increases quadratically following $O(T^2)$ while the cost of FF increases linearly.
Therefore, reducing the MHSA computation is very important for efficient realization of Transformer models.

\subsection{Attention Map Sparsification}\label{ssec:related_attention}

Numerous studies have proposed techniques to selectively compute elements of the attention map~\cite{surveyefficient}.
Patterned attention computation approaches, which select elements in a fixed manner, have been introduced~\cite{sparsetransformer,gpt3,longformer}.
For example, Sparse Transformer~\cite{sparsetransformer} exploits strided pattern that only attends $\ell$ local positions and positions of stride $\ell$, resulting in a $O(T\sqrt{T})$ complexity.
Depending on the pattern, these methods can be supported on modern accelerators with custom kernel implementation.
Adaptive element selection approaches, which dynamically decide elements to compute, have also been studied~\cite{clustertransformer,reformer,routingtransformer}.
For example, Reformer~\cite{reformer} exploits locality-sensitive hashing for clustering so that the attention map can be computed only within the grouped elements.
However, these clustering-based methods require additional computation steps and the access positions dynamically change depending on the input.
The aforementioned studies focus on how to reduce the cost of the attention dot product, while not changing the overall structure of the Transformer model.

\subsection{Removing Attention-related Blocks}\label{ssec:related_remove}

To build an efficient SA mechanism, many studies have focused on the structured removal of a large chunk of computation blocks.
The candidate for the removal (pruning) can be attention heads or attention layers.
For attention head pruning, previous studies have reported that pruning out certain attention heads does not affect the final performance~\cite{michel2019sixteen,voita2019analyzing,metahead}.
Specifically, redundant or less important attention heads in SA can be pruned without degrading the performance for speech recognition~\cite{zhang2020stochastic} and auto-regressive language modeling~\cite{head-pruning}.
Similarly, layer-level pruning has also been studied~\cite{fan2019reducing,sajjad2020poor,dynabert} for natural language processing tasks.
LayerDrop~\cite{fan2019reducing} randomly omits the residual connection during training to make the model more robust to layer pruning.
Our goal is to provide a comprehensive understanding on the impact of the specific structured removal approach, attention map reuse.


\section{Transformer for Speech Recognition}\label{sec:speech}

\subsection{Conformer Architecture}

Transformer-based models have been actively employed for state-of-the-art speech recognition, replacing the previous RNN-based or CNN-based models~\cite{ng2021pushing,zhang2020pushing} thanks to their ability to extract informative features from the input utterance.
In particular, previous works discovered that SA automatically learns to extract useful phonological features~\cite{yang20i_interspeech,attention-reuse} during training.
The input of a Transformer model is a sequence of audio features (frames) extracted by short-time Fourier-transform (STFT), and the output is a sequence of transcribed words.
Because the input and output domains are different, the model internally performs a transformation that turns audio features into text features while passing through a stack of Transformer layers.

\begin{figure*}[h]
    \centering
    \includegraphics[width=0.84\linewidth]{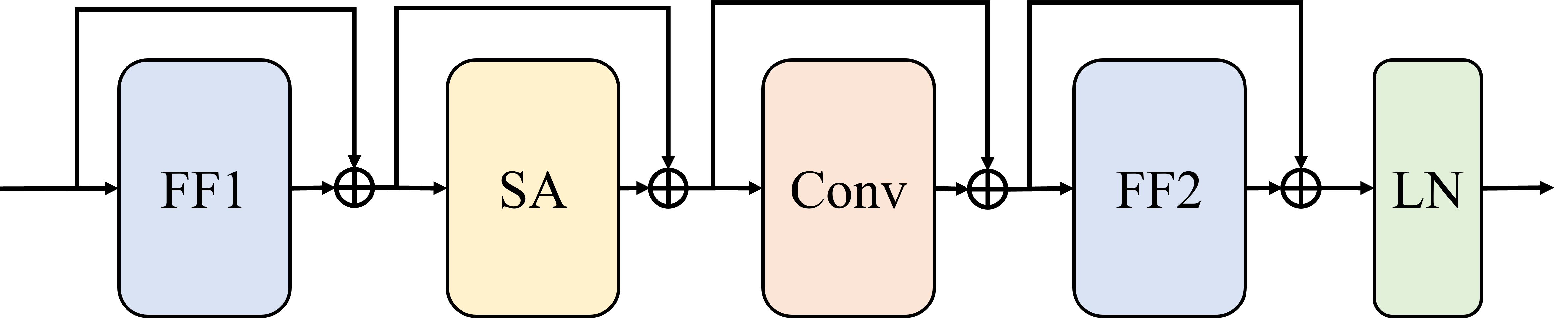}
    \caption{Illustration of a Conformer layer consists of 4 submodules.
    }
    \vspace{-0.5cm}
    \label{fig:conformer}
\end{figure*}

Following the previous work~\cite{attention-reuse}, we employ Conformer~\cite{conformer}, a variant of Transformer widely used for speech recognition.
Conformer consists of 4 submodules as illustrated in Figure~\ref{fig:conformer}.
The main architectural difference between Conformer and Transformer is that Conformer includes two more submodules: an additional FF submodule at the front and the intermediate convolutional (Conv) submodule.
Considering that speech is a continuous signal and nearby frames are highly dependent on each other, the convolutional module is beneficial for enhancing the local relationship between frames that might not be emphasized in SA.
By utilizing both SA and Conv, Conformer achieved a state-of-the-art recognition performance with much fewer parameters than the Transformer-based model without the Conv submodule~\cite{conformer}.

\begin{SCfigure}[][t]
    \vspace{-0.1cm}
    \includegraphics[width=0.30\textwidth]{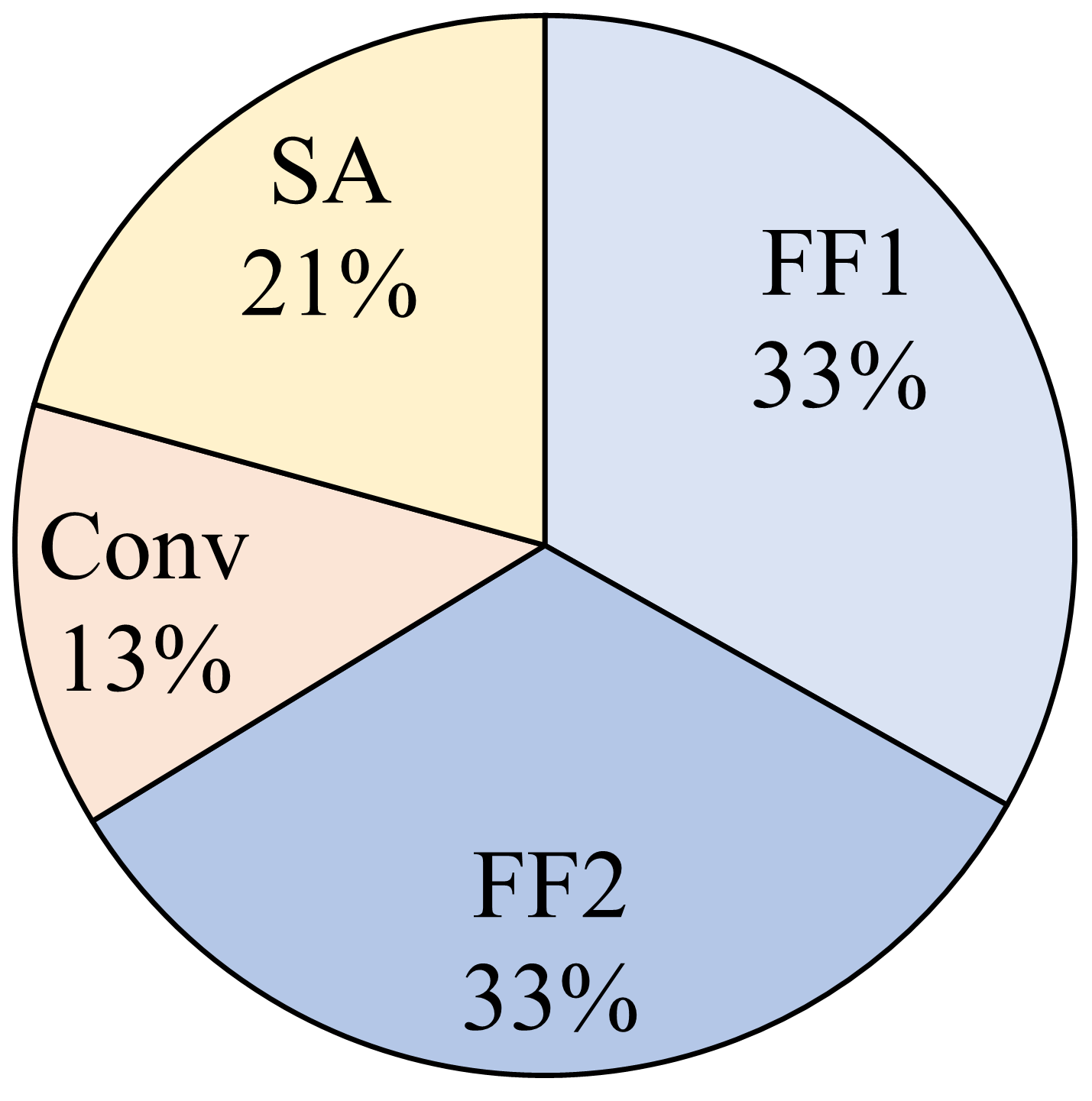}
    \caption{The ratio of the parameter size occupied by each Conformer submodule.
    The values are based on the Conformer-M model.
    }
    \vspace{-0.2cm}
    \label{fig:conformer_params}
\end{SCfigure}

\subsection{Breakdown Analysis}

To understand the bottleneck of Transformer-based models, we analyze how much resource each submodule takes.
We first show the parameter size of each component in Figure~\ref{fig:conformer_params}.
We consider the Conformer-M model, which consists of $L$=16 layers of hidden dimension of $d$=256 and $H$=4 attention heads.
We can observe that about 66\% of parameters come from the FF submodule and SA takes only 21\% of parameters.
The reason for this imbalance is that each FF includes $8d^2$ parameters (therefore, $16d^2$ parameters for two FFs) while SA has $5d^2$ parameters.
In other words, we need to reduce the FF parameters if the target system does not equip enough memory space.
However, the slowest submodule is not FF but SA.

\begin{figure*}[h]
    \centering
    \includegraphics[width=1.0\linewidth]{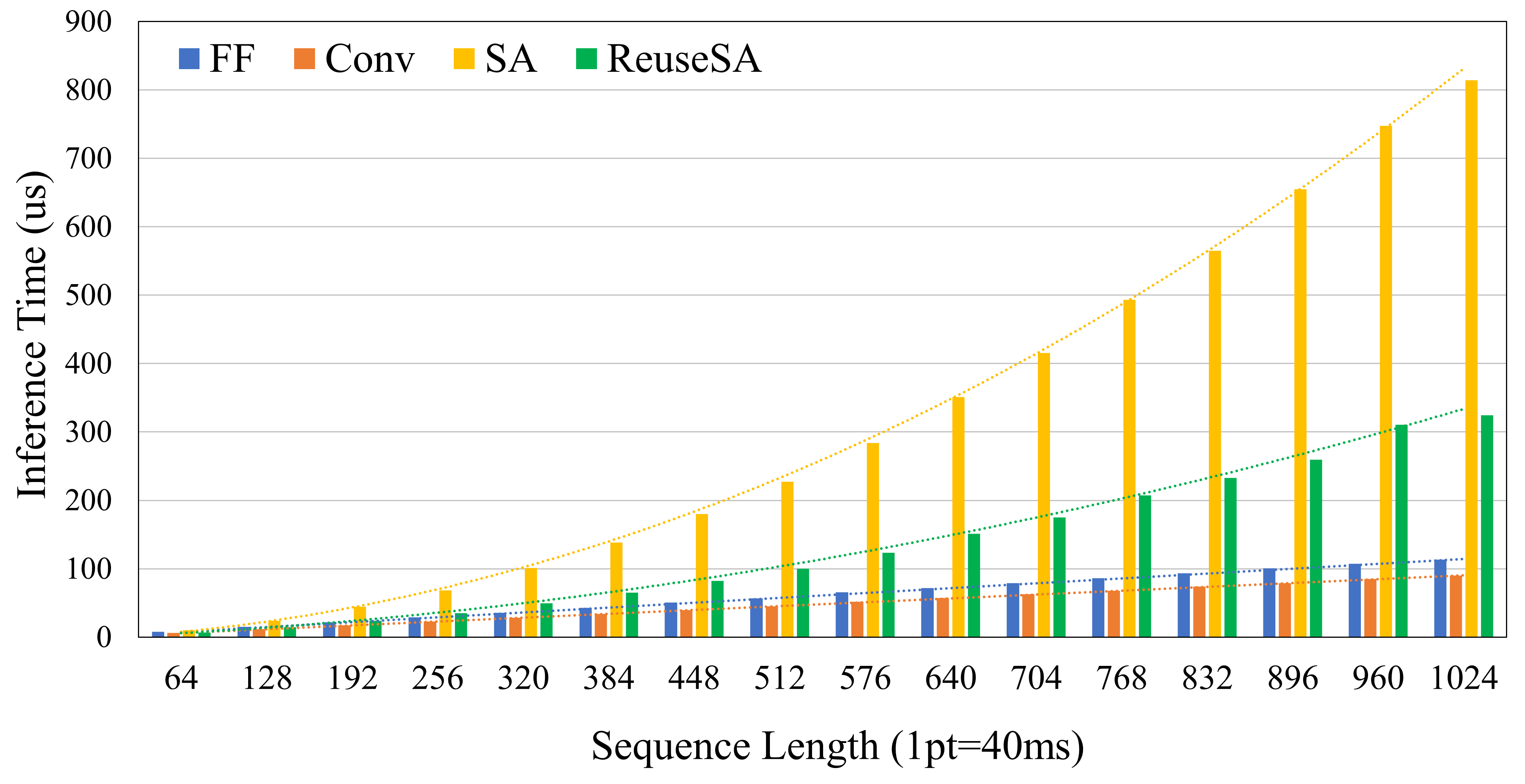}
    \caption{Inference time (us) of each Conformer submodule.
    The order of vertical bars is FF, Conv, SA, and ReuseSA.
    Attention map reuse replaces the SA module with the ReuseSA module and significantly reduces the inference time.
    $x$-axis is the number of frames in the input sequence.
    }
    \vspace{-0.1cm}
    \label{fig:breakdown_conformer}
\end{figure*}

Figure~\ref{fig:breakdown_conformer} presents the inference speed breakdown for different input sequence lengths (we will discuss ReuseSA in the next section).
Note that the $x$-axis of the figure indicates the number of feature frames extracted with a 40ms stride. 
The lengths 256, 512, and 1024 can be reinterpreted to about 10, 20, and 40-second input audio lengths, respectively.
The results are evaluated on a single RTX-Titan GPU, but the tendency should be similar for CPU platforms.
As input length increases, SA requires a quadratic computation cost while FF and Conv only need linearly increasing costs.
Therefore, SA takes 45\% of total inference time for a 10-second input but 67\% for a 30-second input.
This is highly problematic because many speech-related tasks, such as conference transcription, listening comprehension, and speech-to-speech translation, often process much longer utterances than 30-second.

\section{Evaluation of Attention Map Reuse}\label{sec:reusing}


\begin{figure*}[t]
    \centering
    \vspace{0.5cm}
    \includegraphics[width=0.54\linewidth]{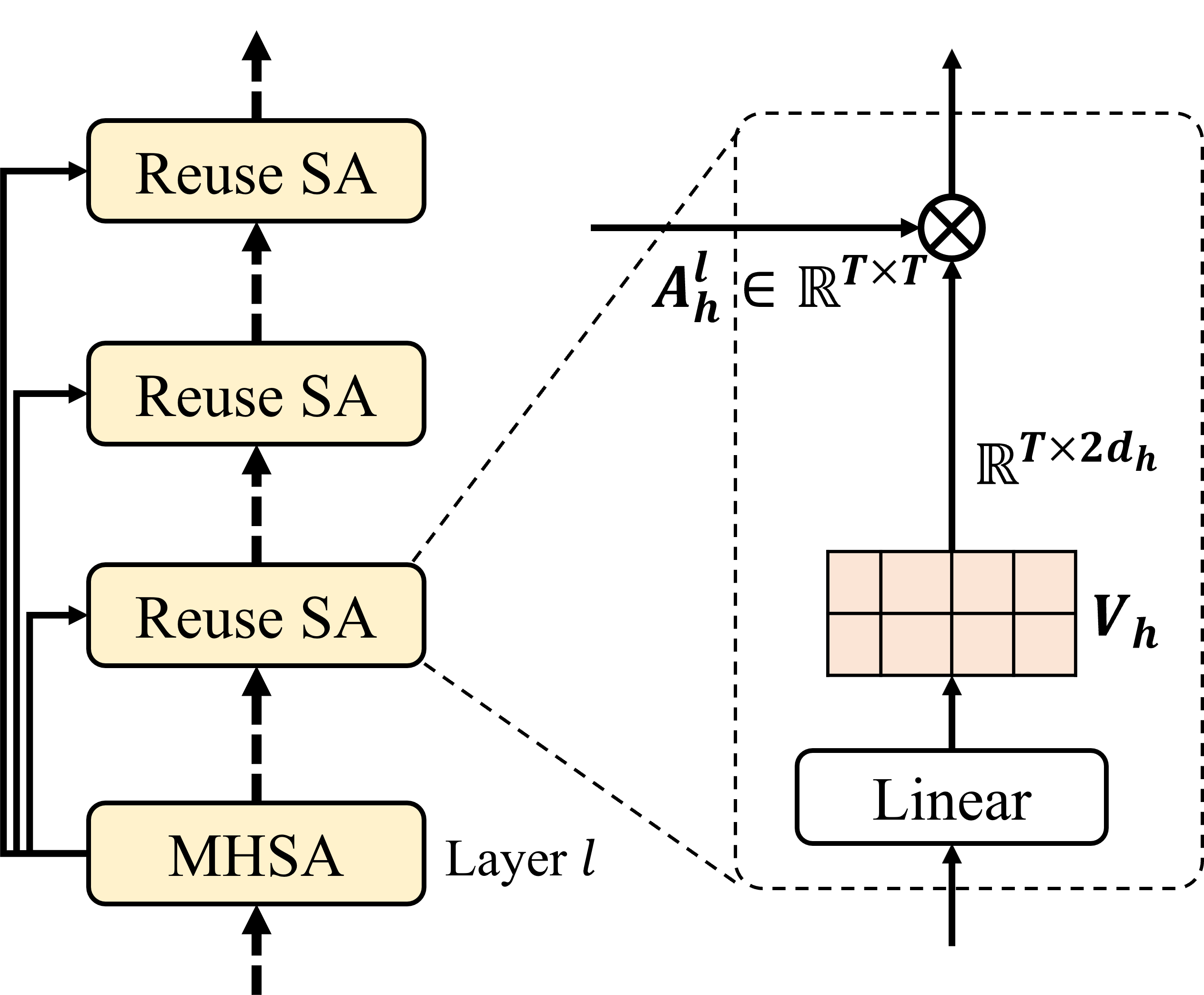}
    \caption{Illustration of each attention head for attention map reuse, in case of the $4\times 4$ reuse configuration.
    The computed attention map $A^{\ell}$ from $\ell\text{-th}$ layer is reused for the next three layers.
    For simplicity, other submodules are omitted.
    }
    \vspace{0.1cm}
    \label{fig:reuse_concept}
\end{figure*}

\subsection{Attention Map Reuse}\label{ssec:reuse_detail}

From the observation that the behavior of SA is very similar between neighboring layers, attention map reuse only computes a single attention map through $M$ consecutive SA layers.
Figure~\ref{fig:reuse_concept} illustrates the attention reuse procedure.
For example, if the attention map from $1^\text{st}$ layer $A^1$ is shared through layers $2\sim M$, the SA output of those layers can be easily computed as below:
\begin{equation}
    \text{SA}^i_h(X^i) = A^1_h V^i_h ,\quad i \in \{2,3, ..., M\} \\
\end{equation}
where $i$ is the layer index and $V^i_h$ is the value computed from each layer.
In other words, attention map reuse groups a set of layers that includes one original SA layer at the front and following $M-1$ reuse SA layers.
The remaining parts of the model, such as FF and Conv submodules, are unchanged.
By omitting the attention map computation, the effective number of SA calculations can be decreased by $M$ times.
Instead of optimizing each SA mechanism, attention map reuse exploits the characteristic of SA and proposes a new axis for Transformer compression.

After applying attention map reuse, the query and key are not needed for the following SA layers and their associated parameters can be removed.
To compensate for the reduced parameter size, previous work suggested increasing the hidden dimension of value~\cite{attention-reuse}.
We follow the same strategy, which makes the size of $W_{V_h}$ from $\mathbb{R}^{T\times d_h}$ to $\mathbb{R}^{T\times 2d_h}$, and the size of $W_O$ from $\mathbb{R}^{d\times d}$ to $\mathbb{R}^{2d\times d}$ (see Figure~\ref{fig:reuse_concept}).
The number of attention heads is preserved.
Note that attention map reuse is not a fine-tuning approach that starts from the fully converged model; we train the modified model from scratch.

\setlength{\tabcolsep}{8pt}
\begin{table*}[t]
    \centering
    \vspace{0.3cm}
    \caption{GPU Inference time (ms) of a speech recognition model with different attention map reuse configurations.
    The numbers from 128 to 1024 indicate the frames in the input utterance.
    Configuration $1\times 16$ represents the baseline model.
    Each frame corresponds to a 40ms stride.
    }
    \resizebox{1.0\linewidth}{!}{
    \begin{tabular}{c|rrrrrrrr}
        \toprule
        Config. & 128 & 256 & 384 & 512 & 640 & 768 & 896 & 1024 \\
        \midrule
        $1\times 16$    & 3.32 & 4.16 & 7.66 & 11.77 & 17.11 & 23.10 & 30.27 & 37.82 \\
        $2\times 8$     & 2.40 & 3.38 & 5.89 & 8.61 & 12.11 & 15.87 & 20.27 & 24.51 \\
        $4\times 4$     & 1.99 & 2.91 & 4.92 & 6.93 & 9.53 & 12.18 & 15.28 & 17.98 \\
        $8\times 2$     & 1.25 & 2.56 & 4.11 & 5.75 & 7.72 & 10.03 & 12.66 & 14.63 \\
        \bottomrule
    \end{tabular}}
    \label{tab:reuse_gpu}
    \vspace{0.2cm}
\end{table*}

\setlength{\tabcolsep}{6pt}
\begin{table*}[t]
    \centering
    \caption{CPU Inference time (ms) of a speech recognition model with different attention map reuse configurations.
    }
    \resizebox{1.0\linewidth}{!}{
    \begin{tabular}{c|rrrrrrrr}
        \toprule
        Config. & 128 & 256 & 384 & 512 & 640 & 768 & 896 & 1024 \\
        \midrule
        $1\times 16$    & 62.97 & 82.93 & 114.44 & 164.48 & 226.88 & 364.05 & 398.36 & 1033.59 \\
        $2\times 8$     & 54.08 & 70.84 & 87.05 & 124.01 & 152.35 & 255.58 & 315.91 & 630.30 \\
        $4\times 4$     & 50.24 & 63.72 & 77.21 & 100.87 & 135.63 & 182.08 & 231.14 & 398.57 \\
        $8\times 2$     & 42.66 & 55.20 & 67.39 & 96.28 & 116.06 & 136.54 & 191.17 & 281.11 \\
        \bottomrule
    \end{tabular}}
    \label{tab:reuse_cpu}
    \vspace{-0.2cm}
\end{table*}
\setlength{\tabcolsep}{4pt}

\subsection{Reuse Analysis}

We analyze the effect of attention map reuse using different configurations.
Tables~\ref{tab:reuse_gpu} and~\ref{tab:reuse_cpu} show the inference time of different reuse configurations.
The configuration $A\times B$ indicates that $A$ layers are grouped to share one attention map and total $B$ groups of layers exist.
The baseline is a 16-layer Conformer-M model~\cite{conformer}, which can be represented as the configuration of $1\times 16$.
The results are measured on an RTX Titan GPU and an Intel Xeon Gold 6130 CPU.
We evaluate three different attention map reuse configurations, $2\times 8$, $4\times 4$, and $8\times 2$ where the number of attention map computations is 8, 4, and 2, respectively.
Although we can combine different numbers of layers to form heterogeneous groups, we find that the number of unique attention map computations determines the total latency of the model.
Note that there is a trade-off between the performance and the inference speed if the number of unique attention map computations is too small (see  Section~\ref{sssec:performance}).

We observe a clear improvement in the inference speed for both GPU and CPU as more layers are grouped.
For $4\times 4$ configuration, 10-second (length 256), 20-second (length 512) and 40-second (length 1024) utterance saves about 38\%, 41\%, and 52\% of the GPU inference time, respectively.
The same configuration saves about 33\%, 39\%, and 61\% for the CPU inference time.
CPU inference time gain is not as good as that of GPU when the length is short but provides a higher benefit when the length is long.
Figure~\ref{fig:breakdown_conformer} also demonstrates the effect of reuse in the case of $4\times 4$.
Comparing the original SA and the reuse SA, the reuse SA takes about 35\% to 45\% of the latency of the original SA.

\subsection{Discussion}

\subsubsection{Attention Map Reuse in Natural Language Processing}
Attention map reuse was applied for neural machine translation~\cite{xiao2019sharing} and BERT-based language modeling~\cite{lazyformer}.
For machine translation, the average speedup was about 1.3 times without considerable performance degradation~\cite{xiao2019sharing}.
The inference speedup is less than in speech recognition because machine translation usually considers a relatively shorter sequence length $T$ (shorter than 50) and larger hidden dimension $d$.
If $d$ is much larger than $T$, the $O(T^2)$ SA computation does not dominate the total inference cost, so the advantage from attention map reuse is limited.
For the language model inference, the speedup was about 1.3 times with a marginal performance gain~\cite{lazyformer}.
However, the work mainly focused on the $2\times 6$ configuration because the more aggressive reuse configuration did not achieve satisfactory performance for GLUE downstream tasks.
We conclude that the efficiency of attention map reuse can be maximized when 1) the expected input sequence length is longer than the hidden dimension, and 2) the target task is not very sensitive to a small attention map variation.
Speech recognition well fits these conditions because it handles very long sequences and the frame features change slowly along the time axis.

\setlength{\tabcolsep}{5pt}
\begin{table*}[t]
    \centering
    \caption{Word error rate (\%) for different attention map reuse configurations.}
    \resizebox{0.94\textwidth}{!}{
    \begin{tabular}{c|c|cccc}
        \toprule
        Config. & \#Param (M) & \textit{dev-clean} & \textit{dev-other} & \textit{test-clean} & \textit{test-other} \\
        \midrule
        $1\times16$  &25.45  & 3.1  & 8.3  & \textbf{3.2}      & 8.4 \\
        $2\times 8$  &24.92  & \textbf{3.0}  & \textbf{8.2}  & 3.3      & \textbf{8.2} \\
        $4\times 4$  &24.66  & \textbf{3.0}  & \textbf{8.2}  & 3.3      & \textbf{8.2} \\
        $8\times 2$  &24.52  & 3.3  & 8.8  & 3.6      & 8.7 \\
        \bottomrule
    \end{tabular}
    }
    \label{tab:reuse_result}
    \vspace{-0.3cm}
\end{table*}
\setlength{\tabcolsep}{4pt}

\subsubsection{Effect on Performance}\label{sssec:performance}
In Table~\ref{tab:reuse_result}, we show the word error rate (WER) of attention map reuse on LibriSpeech~\cite{librispeech} speech recognition dataset, which includes four evaluation data subsets\footnote{Four LibriSpeech subsets are \textit{dev-clean}, \textit{dev-other}, \textit{test-clean}, and \textit{test-other}. The postfix `other' means that the subset is more challenging.}.
We borrow the result from our original paper~\cite{attention-reuse} on speech recognition to briefly show that attention map reuse can be employed without affecting the performance.
In short, recognition performance is almost preserved for $2\times 8$ and $4\times 4$ configurations but not for the $8\times 2$ case.
The authors suggested that the capacity may become insufficient to internalize the necessary information for every layer in a group when too many layers share the same attention map.
Note that the number of parameters is almost the same for configurations because we doubled the dimension of value (see Section~\ref{ssec:reuse_detail}).

\subsubsection{Attention Computation Reduction in Speech Recognition}
Many works have been proposed techniques to reduce the computational cost of SA, especially for speech recognition.
Local windowing is the common approach that only exploits a limited range of frames for attention map computation.
For example, each frame may only consider neighboring frames (e.g., only accessing past 64 and future 64 frames) as candidates of the attention mechanism; this approach decreases the complexity of SA from $O(T^2)$ to $O(TR)$, where $R$ is the number of accessible frames.
However, restricting the range often lowers the recognition performance over full sequence-based models~\cite{dualmode,streaminggap}.
On the other hand, several studies designed more efficient Transformer models for speech recognition~\cite{efficient-conformer,simplified-sa}.
These approaches, including faster query-key dot product~\cite{simplified-sa} and time-strided SA~\cite{efficient-conformer}, are orthogonal to attention map reuse and can be used together.

\section{Conclusion}\label{sec:conclusion}

In this paper, we analyzed a recently proposed efficient SA compression method, named attention map reuse, for Transformer-based speech recognition.
We first perform a detailed analysis of the inference bottleneck of the Conformer model used for speech processing, evaluated on a wide range of input sequence lengths.
Our analysis provides a thorough understanding on the burden of SA when using Transformer in practice.
Then, we demonstrate the computational savings from attention map reuse on GPU and CPU platforms.
We claim that attention map reuse is a very promising method for utilizing Transformer-based models on modern hardware systems.


\bibliographystyle{splncs04}
\bibliography{reference}

\begin{thebibliography}{10}
\providecommand{\url}[1]{\texttt{#1}}
\providecommand{\urlprefix}{URL }
\providecommand{\doi}[1]{https://doi.org/#1}

\bibitem{layernorm}
Ba, J.L., Kiros, J.R., Hinton, G.E.: Layer normalization. arXiv preprint
  arXiv:1607.06450  (2016)

\bibitem{longformer}
Beltagy, I., Peters, M.E., Cohan, A.: Longformer: The long-document
  transformer. arXiv preprint arXiv:2004.05150  (2020)

\bibitem{gpt3}
Brown, T., Mann, B., Ryder, N., Subbiah, M., Kaplan, J.D., Dhariwal, P.,
  Neelakantan, A., Shyam, P., Sastry, G., Askell, A., et~al.: Language models
  are few-shot learners. Advances in neural information processing systems
  \textbf{33},  1877--1901 (2020)

\bibitem{efficient-conformer}
Burchi, M., Vielzeuf, V.: Efficient conformer: Progressive downsampling and
  grouped attention for automatic speech recognition. arXiv preprint
  arXiv:2109.01163  (2021)

\bibitem{sparsetransformer}
Child, R., Gray, S., Radford, A., Sutskever, I.: Generating long sequences with
  sparse transformers. arXiv preprint arXiv:1904.10509  (2019)

\bibitem{performer}
Choromanski, K.M., Likhosherstov, V., Dohan, D., Song, X., Gane, A., Sarlos,
  T., Hawkins, P., Davis, J.Q., Mohiuddin, A., Kaiser, L., et~al.: Rethinking
  attention with performers. In: International Conference on Learning
  Representations (2021)

\bibitem{streaminggap}
Doutre, T., Han, W., Chiu, C.C., Pang, R., Siohan, O., Cao, L.: {Bridging the
  Gap Between Streaming and Non-Streaming ASR Systems by Distilling Ensembles
  of CTC and RNN-T Models}. In: Proc. Interspeech 2021. pp. 1807--1811 (2021).
  \doi{10.21437/Interspeech.2021-637}

\bibitem{fan2019reducing}
Fan, A., Grave, E., Joulin, A.: Reducing transformer depth on demand with
  structured dropout. In: International Conference on Learning Representations
  (2019)

\bibitem{conformer}
Gulati, A., Qin, J., Chiu, C.C., Parmar, N., Zhang, Y., Yu, J., Han, W., Wang,
  S., Zhang, Z., Wu, Y., Pang, R.: Conformer: Convolution-augmented transformer
  for speech recognition. In: Proc. Interspeech 2020. pp. 5036--5040 (2020)

\bibitem{lstm}
Hochreiter, S., Schmidhuber, J.: Long short-term memory. Neural computation
  \textbf{9}(8),  1735--1780 (1997)

\bibitem{dynabert}
Hou, L., Huang, Z., Shang, L., Jiang, X., Chen, X., Liu, Q.: Dynabert: Dynamic
  bert with adaptive width and depth. Advances in Neural Information Processing
  Systems  \textbf{33},  9782--9793 (2020)

\bibitem{reformer}
Kitaev, N., Kaiser, L., Levskaya, A.: Reformer: The efficient transformer. In:
  International Conference on Learning Representations (2019)

\bibitem{li2020efficient}
Li, B., Kong, Z., Zhang, T., Li, J., Li, Z., Liu, H., Ding, C.: Efficient
  transformer-based large scale language representations using
  hardware-friendly block structured pruning. In: Findings of the Association
  for Computational Linguistics: EMNLP 2020. pp. 3187--3199 (2020)

\bibitem{simplified-sa}
Luo, H., Zhang, S., Lei, M., Xie, L.: Simplified self-attention for
  transformer-based end-to-end speech recognition. In: 2021 IEEE Spoken
  Language Technology Workshop (SLT). pp. 75--81. IEEE (2021)

\bibitem{michel2019sixteen}
Michel, P., Levy, O., Neubig, G.: Are sixteen heads really better than one?
  Advances in Neural Information Processing Systems  \textbf{32},  14014--14024
  (2019)

\bibitem{ng2021pushing}
Ng, E.G., Chiu, C.C., Zhang, Y., Chan, W.: Pushing the limits of
  non-autoregressive speech recognition. arXiv preprint arXiv:2104.03416
  (2021)

\bibitem{librispeech}
Panayotov, V., Chen, G., Povey, D., Khudanpur, S.: Librispeech: an asr corpus
  based on public domain audio books. In: 2015 IEEE international conference on
  acoustics, speech and signal processing (ICASSP). pp. 5206--5210. IEEE (2015)

\bibitem{qi2021accelerating}
Qi, P., Sha, E.H.M., Zhuge, Q., Peng, H., Huang, S., Kong, Z., Song, Y., Li,
  B.: Accelerating framework of transformer by hardware design and model
  compression co-optimization. In: 2021 IEEE/ACM International Conference On
  Computer Aided Design (ICCAD). pp.~1--9. IEEE (2021)

\bibitem{routingtransformer}
Roy, A., Saffar, M., Vaswani, A., Grangier, D.: Efficient content-based sparse
  attention with routing transformers. Transactions of the Association for
  Computational Linguistics  \textbf{9},  53--68 (2021)

\bibitem{sajjad2020poor}
Sajjad, H., Dalvi, F., Durrani, N., Nakov, P.: Poor man’s bert: Smaller and
  faster transformer models. arXiv preprint arXiv:2004.03844  (2020)

\bibitem{head-pruning}
Shim, K., Choi, I., Sung, W., Choi, J.: Layer-wise pruning of transformer
  attention heads for efficient language modeling. In: 2021 18th International
  SoC Design Conference (ISOCC). pp. 357--358. IEEE (2021)

\bibitem{attention-reuse}
Shim, K., Choi, J., Sung, W.: Understanding the role of self attention for
  efficient speech recognition. In: International Conference on Learning
  Representations (2022)

\bibitem{surveyefficient}
Tay, Y., Dehghani, M., Bahri, D., Metzler, D.: Efficient transformers: A
  survey. arXiv preprint arXiv:2009.06732  (2020)

\bibitem{transformer}
Vaswani, A., Shazeer, N., Parmar, N., Uszkoreit, J., Jones, L., Gomez, A.N.,
  Kaiser, {\L}., Polosukhin, I.: Attention is all you need. Advances in neural
  information processing systems  \textbf{30} (2017)

\bibitem{voita2019analyzing}
Voita, E., Talbot, D., Moiseev, F., Sennrich, R., Titov, I.: Analyzing
  multi-head self-attention: Specialized heads do the heavy lifting, the rest
  can be pruned. In: Proceedings of the 57th Annual Meeting of the Association
  for Computational Linguistics. pp. 5797--5808 (2019)

\bibitem{clustertransformer}
Vyas, A., Katharopoulos, A., Fleuret, F.: Fast transformers with clustered
  attention. Advances in Neural Information Processing Systems  \textbf{33},
  21665--21674 (2020)

\bibitem{linformer}
Wang, S., Li, B.Z., Khabsa, M., Fang, H., Ma, H.: Linformer: Self-attention
  with linear complexity. arXiv preprint arXiv:2006.04768  (2020)

\bibitem{xiao2019sharing}
Xiao, T., Li, Y., Zhu, J., Yu, Z., Liu, T.: Sharing attention weights for fast
  transformer. In: Proceedings of the Twenty-Eighth International Joint
  Conference on Artificial Intelligence (IJCAI-19) (2019)

\bibitem{yang20i_interspeech}
Yang, S., Liu, A.T., yi~Lee, H.: Understanding self-attention of
  self-supervised audio transformers. In: Proc. Interspeech 2020. pp.
  3785--3789 (2020). \doi{10.21437/Interspeech.2020-2231}

\bibitem{lazyformer}
Ying, C., Ke, G., He, D., Liu, T.Y.: Lazyformer: Self attention with lazy
  update. arXiv preprint arXiv:2102.12702  (2021)

\bibitem{dualmode}
Yu, J., Han, W., Gulati, A., Chiu, C.C., Li, B., Sainath, T.N., Wu, Y., Pang,
  R.: Dual-mode asr: Unify and improve streaming asr with full-context
  modeling. In: International Conference on Learning Representations (2021)

\bibitem{zhang2020stochastic}
Zhang, S., Loweimi, E., Bell, P., Renals, S.: Stochastic attention head
  removal: A simple and effective method for improving transformer based asr
  models. arXiv preprint arXiv:2011.04004  (2020)

\bibitem{zhang2020pushing}
Zhang, Y., Qin, J., Park, D.S., Han, W., Chiu, C.C., Pang, R., Le, Q.V., Wu,
  Y.: Pushing the limits of semi-supervised learning for automatic speech
  recognition. arXiv preprint arXiv:2010.10504  (2020)

\bibitem{metahead}
Zhang, Z., Qi, F., Liu, Z., Liu, Q., Sun, M.: Know what you don't need:
  Single-shot meta-pruning for attention heads. AI Open  \textbf{2},  36--42
  (2021)

\end{thebibliography}

\end{document}